\PassOptionsToPackage{pdfencoding=auto, psdextra}{hyperref}
\documentclass[3p,10pt,number]{elsarticle}
\usepackage{placeins}
\usepackage{amssymb}
\usepackage{amsmath}
\usepackage{amsthm}
\usepackage{xcolor}
\usepackage{graphicx}
\usepackage{hyperref}
\hypersetup{
    pdfencoding=auto,
    psdextra,
    colorlinks=true,
    linkcolor=blue,
    citecolor=blue,
    urlcolor=blue,
    pdfauthor={Maryam SaniSales, Ali Vefghi, Zahed Rahmati, Ali Darvishi},
    pdftitle={Dust Source Emmision Forecasting},
    pdfsubject={Subject},
    pdfkeywords={key1, key2, key3}
}
\usepackage{multirow}

\usepackage{url}
\usepackage{booktabs} % For better horizontal rules in tables
\usepackage{siunitx}  % For aligning numbers by decimal point
\usepackage{caption}  % For table captions
\usepackage{xcolor}
\usepackage{makecell}
\usepackage[table]{xcolor}

\usepackage[pdfencoding=auto, psdextra]{hyperref}
\journal{a Journal}

\usepackage{lineno}
\begin{document}
\begin{frontmatter}

\title{Enhancing Graph Neural Networks Using Proximity Graphs for Dust Source Emission Forecasting}

		\author[label1]{Maryam Sanisales} \ead{m.sanisales@aut.ac.ir}
		\author[label1]{Zahed Rahmati\corref{cor1}} 
        \ead{zrahmati@aut.ac.ir}
		\author[label2]{Ali Darvishi Boloorani} \ead{ali.darvishi@ut.ac.ir}
        \author[label1]{Ali Vefghi} 
        \ead{alivefghi@aut.ac.ir}
		
		\affiliation[label1]{organization={Department of Mathematics and Computer Science,  Amirkabir University of Technology},
			city={Tehran},
			country={Iran}}
		\affiliation[label2]{organization={Department of Remote Sensing and GIS, Faculty of Geography, University of Tehran}, 
			city={Tehran},
			country={Iran}}	
		
		\cortext[cor1]{Corresponding author}
				
		\author{} %% Author name
%% Abstract
\begin{abstract}
Accurate prediction of dust source emissions is critical for mitigating the significant environmental and health hazards posed by dust storms. Traditional forecasting methods often struggle to capture the complex spatiotemporal dynamics of these phenomena.

In this paper, we demonstrate that proximity graphs enable Graph Neural Networks (GNNs) to effectively model the intricate spatial and temporal relationships between data points. Specifically, we use proximity graphs—such as Delaunay triangulation, Gabriel graph, $k$-Nearest Neighbor graph, and Yao graph—as the input for GNNs (including GraphSAGE, Graph Convolutional Networks, and Graph Attention Networks) to perform message passing.

Our approach highlights the effectiveness of integrating proximity graphs with GNNs for robust and accurate dust source forecasting. To emphasize the importance of proximity graph representations, we compare our method against GNNs using random graphs for message passing. The results show that GNNs with proximity graphs significantly outperform those with random graphs and are also far superior to Long Short-Term Memory (LSTM) model in dust source emission forecasting.
\end{abstract}

% \includegraphics[width=\textwidth]{g_abstract.pdf}


%% Keywords
\end{frontmatter}

\section{Introduction}

Dust storms are one of the most significant forms of atmospheric pollution, posing serious environmental challenges to ecosystems, air quality, health, and climate stability. They consist of particulate matter-mainly small grains of loose soil and sand, that can be carried by surface winds and transported thousands of kilometers beyond their sources~\cite{1,4}. Large-scale weather patterns drive this long-range dust transport, making desert dust a major contributor to climate change. In addition to its climatic impact, dust is harmful to human health, causing respiratory and cardiovascular diseases~\cite{2,3}, while also disrupting agriculture, transportation, and energy production, such as reducing the efficiency of solar panels~\cite {1a}. Effective management of dust storms is therefore crucial for protecting infrastructure and public health.

This study focuses on the Tigris and Euphrates river basin, a region where dust events are frequent, with a peak season from spring to summer. It is expected that climate change will result in elevated temperatures and diminished precipitation, leading to the expansion of arid regions~\cite{9b,2b}. Consequently, the occurrence of dust storms is anticipated to intensify. These intense dust storms will have greater impacts on the weather~\cite{30b}.

Accurate forecasting of dust source emissions is vital for policymakers, businesses, and citizens to plan mitigation measures. However, making reliable predictions requires high-quality data. While field measurements offer detailed information, they are impractical for large-scale studies. Remote sensing, though cost-effective and capable of covering vast areas, often suffers from limitations in spatial and temporal resolution~\cite{2a,6,7}. These challenges highlight the need for advanced modeling techniques.

Predicting dust emissions is inherently complex due to their spatiotemporal nature and dependence on multiple environmental factors, such as vegetation cover, wind speed, soil moisture, temperature, precipitation, and topography~\cite{3a,4a,14,15,16}. Traditional machine learning models, including support vector machines, decision trees, and linear regression, have been applied to model spatiotemporal data~\cite{17b,56b,60b}. Still, they often fail to capture the intricate relationships among variables or to effectively learn time-dependent patterns. Deep neural networks, on the other hand, have shown superior capability in recognizing complex patterns in dynamic data~\cite{75b}.

Recent studies have attempted to identify dust hotspots using data mining, ensemble models, and multi-criteria decision-making approaches~\cite{9,10,12,13}. While these methods offer useful insights, they still struggle to fully capture the spatial and temporal interactions underlying dust source dynamics. Therefore, there is a clear need for models that can simultaneously handle both spatial and temporal dependencies.

Graph Neural Networks (GNNs) have emerged as powerful tools for modeling spatio-temporal data due to their ability to represent entities as nodes and their relationships as edges, enabling efficient message passing and pattern learning~\cite{5a,6a}. This study leverages this strength by modeling multiple time steps of dust source emissions as proximity graphs, where nodes correspond to dust sources and edges encode their spatial and temporal interactions. Specifically, we explore various proximity graph constructions (Delaunay, Gabriel, $k$-Nearest Neighbor, and Yao graphs) combined with state-of-the-art static GNN architectures (GraphSAGE, GCN, and GAT). This approach enables us to capture complex spatio-temporal dependencies accurately and predict dust source activity under varying environmental and weather conditions.
To the best of our knowledge, this is the first study to apply proximity graph models with GNNs to dust emission forecasting. Our framework demonstrates significant performance improvements over traditional methods and offers a robust solution for proactive planning and mitigation strategies.

The remainder of this paper is organized as follows. Section 2 describes the study area, dataset, and environmental factors. Section 3 presents our methodology, including graph construction and GNN models. Section 4 details the experimental design, evaluation metrics, and results. Finally, Section 5 summarizes the findings and outlines future research directions.

\section{Methodology}

\begin{figure}[ht]
  \centering
  \includegraphics[width=\columnwidth]{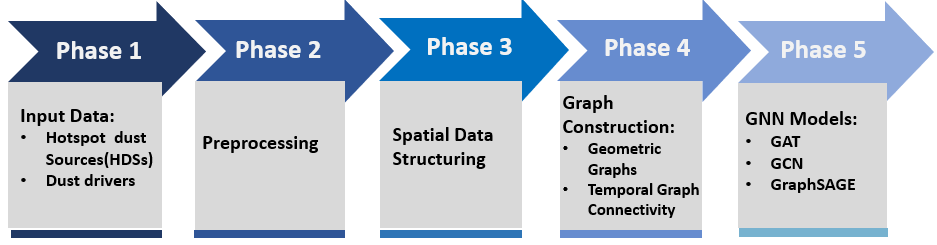}
  \caption{Overview of the proposed pipeline.}
  \label{fig1}
\end{figure}

The proposed methodology for forecasting dust emissions is visually summarized in Figure \ref{fig1}. The illustration outlines a comprehensive pipeline that addresses the challenge as a binary classification problem, with the objective of identifying future active dust source emission based on their historical behaviors.
The approach, when viewed from a high-level perspective, comprises multiple essential components. The following elements are incorporated into the data preprocessing stage: a Spatial Data Structure for the management of spatial data, a graph constructor for the construction of relational structures, a temporal graph connector for the incorporation of time-series dynamics, and finally, GNN models for the predictive task.

\subsection{Data Preparation}

The domain of interest is represented by a collection of spatial-temporal tuples: (x,y,t), where (x,y) is the spatial coordinates and t  denotes the time. A subset of these tuples, designated as A, has been identified as "active hotspots", with a total of 11,618 instances identified. The fundamental characteristic of the system under consideration is as follows: If a specific spatial location, designated by the triples ($x_j$,$y_j$) is active at a given time, denoted by $t_k$, then, for any other time $t_l$ that is not equal to $t_k$, the tuple ($x_j$,$y_j$,$t_k$) is, by definition, "not active". This results in an overwhelming majority of "not active" instances compared to "active" ones across the total 2,616,768 potential combinations of the form: (x,y,t). This presents a substantial class imbalance problem for pattern learning.

The objective is to organize the spatial information into a format that is suitable for the subsequent steps, such as graph construction.  The total dataset for 22 years consists of approximately 11,000 unique hotspots, each monitored on a monthly basis  to capture their spatiotemporal behavior. After transforming this  data into a suitable format, it is divided into monthly intervals, resulting in a set of approximately 11,000 points for each month. To manage the computational cost for the machine learning algorithms, a space sampling procedure is subsequently applied.

A main issue with the data related to the imbalanced distribution of points across months, attributable to the disproportionate number of positive samples relative to negative samples. Addressing this issue is imperative to ensure the integrity and reliability of the data. The utilization of a space sampler serves to diminish the superfluous information present in each month, thereby enhancing the efficiency of data processing. The utilization of space sampling techniques has been demonstrated to effectively mitigate the number of negative samples observed in each month, thereby enhancing the overall reliability of the data. The prevailing approach is random sampling; however, alternative effective sampling techniques may also be employed.

To address the challenge posed by the extensive negative sample space and to optimize computational efficiency, a targeted spatial sampling strategy was implemented. This involved utilizing a nearest neighbor-based approach to construct the negative sample set. Specifically, negative instances were identified and included in the dataset by considering their proximity to active dust source, with the number of $k$-Nearest neighbors explored for this construction set at value of 1.

\subsection{Proximity Graphs}

Graph Neural Networks (GNN) have seen a surge in popularity in recent years due to their capacity to model complex relationships between data points. However, it is crucial to note that the initiation of operations with GNN models is contingent upon the presence of a graph structure in the data. This issue represents a significant challenge for GNN, given that all the data utilized in this study and remote sensing in general are in a tabular structure. A key point to emphasize is that prior to the initiation of the training of GNN models on the dataset, it is imperative to ascertain the underlying relationships between the data points.

In the tabular dataset under consideration, the data is arranged in rows and columns. In the event that a matrix is established for each month, with each row representing a dust emission source and its associated spatial coordinates, and with each column representing the features related to the dust source, which consists of 11 features, this would consequently result in 11 columns.

The objective of this section is to construct homogeneous graphs of the data points for each month, thereby demonstrating the spatial and inter-point relationships through a deliberate graph structure. First, it is necessary to define what a graph is. A homogeneous graph is denoted by $G$. A data graph is defined by nodes and the relationships (edges/links) between different nodes, which are denoted by $(V,E)$, respectively. Accordingly, within the matrix structure, each row is representative of a node (i.e., a dust source emission ), and the edges between them are indicative of their relationships.

In the preceding section, an explanation was provided regarding the graph structure for data. The present section will address the connections between points. In this study, several proximity graph methods were employed to create these edges, which will be discussed in the next sections.

\subsubsection{Delaunay Triangulation}

The initial approach involves the utilization of the Delaunay graph. The delineation of the edges between nodes is derived from a proximity graph known as Delaunay triangulation~\cite{9a}. In this method, the points of a triangle are selected in such a manner that if we pass through those three points of a circle, no other point is located inside that circle. This condition, while seemingly elementary, leads to a noteworthy phenomenon: the interior angles of the triangles approach a maximum value, and the triangles' shape approaches an equilateral one. Additionally, this triangulation process reduces the angles of the triangles to their maximum possible value, thereby preventing the formation of narrow triangles. This function is typically employed to identify the nearest neighbors to a given node~\cite{9a}.

\subsubsection{$k$-Nearest-Neighbor Graph}

The $k$-Nearest Neighbor ($k$-NN) Graph is constructed by connecting each point to its $k$ nearest points.
$k$-NN algorithm functions similarly to a voting system. The classification of a new data point is determined by examining the $k$ nearest points in the data set~\cite{10a}. In this study, we will set $k$ to 1, 2, and 3.

\subsubsection{Gabriel Graph}

A Gabriel graph, a concept in computational geometry and adjacency graphs, establishes a specific relationship between a set of points in Euclidean space. The graph is formally constructed by creating an edge between any two distinct points if and only if the closed disk whose line segment connecting the two points forms its diameter does not contain any other points from the given set~\cite{11a}.

The "empty disk" criterion guarantees that points that are sufficiently "close" to each other are connected, without any intervening points in their immediate vicinity. Consequently, the Gabriel graph offers a visual representation of the concept of local neighborhood based on geometric emptiness. The graph under consideration is a planar graph, which is defined as a graph that can be drawn without any edge intersections. It is known that the graph is a subgraph of the Delaunay triangulation. This property states that all edges in a Gabriel graph are also present in the Delaunay triangulation of the same set of points, thereby offering a sparser representation of adjacency relations.

\subsubsection{Yao Graph}

\begin{figure}[h!]
  \centering
  \includegraphics[width=\columnwidth]{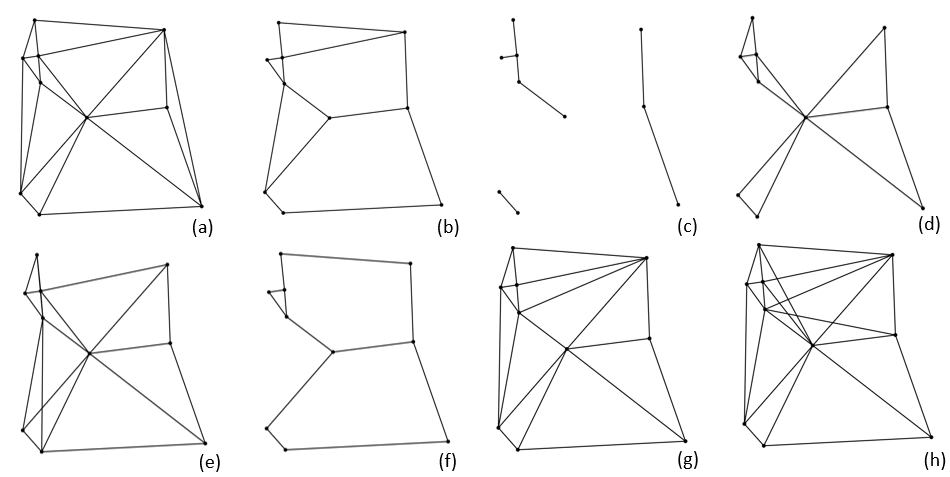}
  \caption{Drawing Proximity Graphs on 10 Nodes: (a) Delaunay Triangulation, (b) Gabriel Graph, (c) 1-Nearest Neighbor (1-NN) Graph, (d) 2-Nearest Neighbor (2-NN) Graph, (e) 3-Nearest Neighbor (3-NN) Graph, (f) $Y_2$  Graph,(g) $Y_4$  Graph, (h) $Y_6$  Graph.}
  \label{fig2}
\end{figure}

In the field of computational geometry, a Yao graph is defined as a specific type of geometric tree. This geometric tree is characterized as an undirected weighted graph that establishes a connection between a set of geometric points~\cite{13a}. The method under consideration involves the division of the spatial domain surrounding each point into subdivisions, with the utilization of rays that are positioned at uniform intervals. Subsequently, each point is linked to its proximate neighbor within each segment. The parameter "r" is instrumental in determining the number of rays and segments, with higher values of "r" resulting in a more precise approximation of the Euclidean distance. The Yao graphs are a special case of graphs that minimize the Euclidean distance between a point and its neighbor. The fundamental concept involves the division of the spatial domain surrounding each point into a predetermined number of equal-angle cones (sectors).  The Yao graph is denoted as $Y_r$  Graph that $r:2$ and 6.
As illustrated in Figure \ref{fig2}, the following examples demonstrate proximity graph constructions for 10 nodes.

Graph Neural Networks (GNNs) operate under two primary learning models: Inductive learning and transductive learning~\cite{8a}. With inductive learning, training and message passing are done between only the training nodes in graph, and test nodes are apart from the training nodes. Conversely, transductive learning allows the training and test nodes to message-pass, and the test labels are not involved during the training phase. Given the inherent nature of dust source emission forecasting, an inductive learning approach is essential. This approach necessitates a model capable of robust generalization to new observations.
\begin{figure}
 \includegraphics[width=.8\columnwidth]{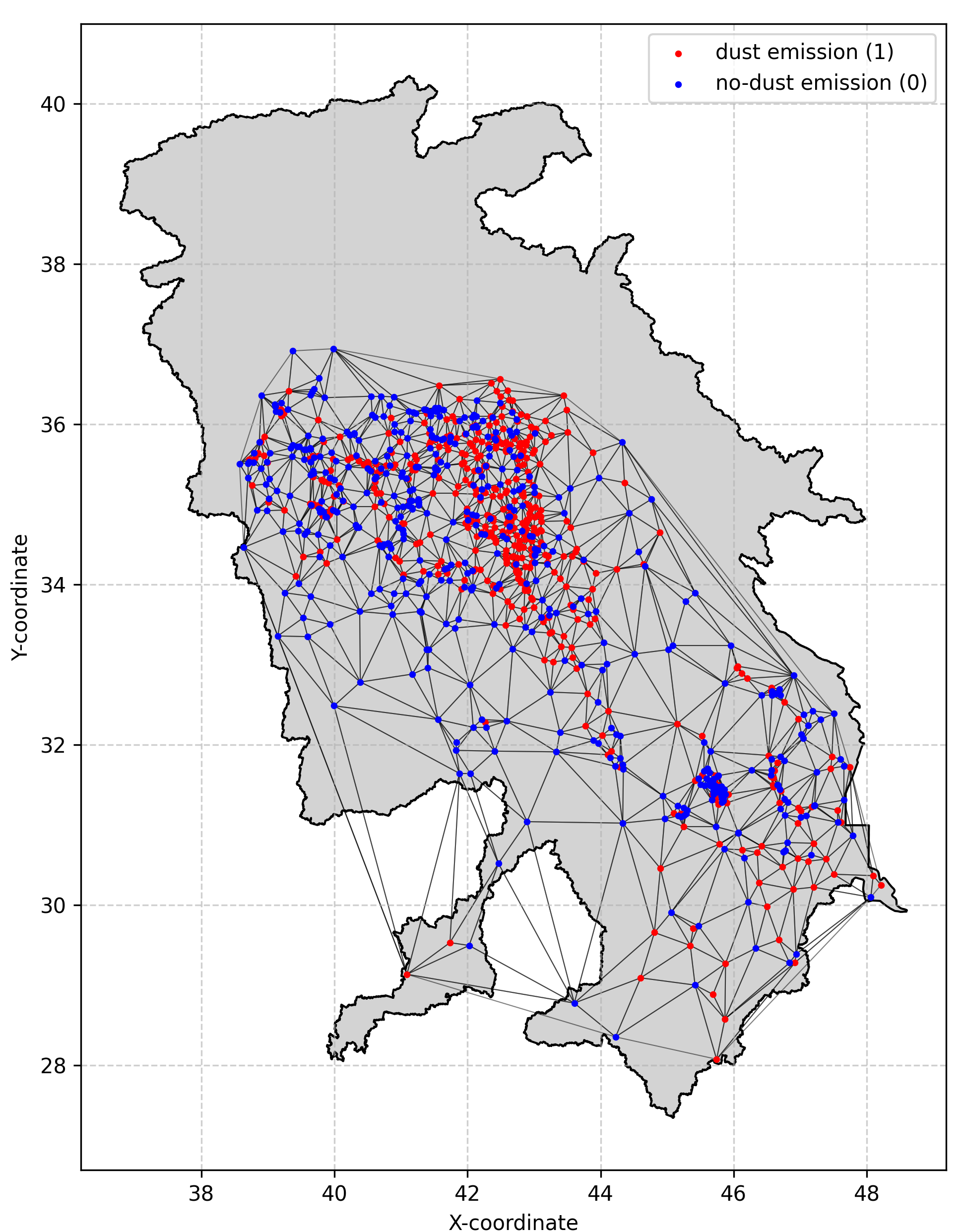}
  \caption{A spatial graph showing the distribution of dust source emission (red points) and non-dust source emission (blue points) within a TEB. The connections represent a Delaunay triangulation of the data points for a single time frame.}
 \label{fig3}
\end{figure}

Figure \ref{fig3} presents a visual representation of the geographical distribution of dust source emissions for a designated month. The red points indicate active "dust emission" sources, which are concentrated in the central and northern parts of the basin, and show the blue points ("no-dust emission"). The creation of edges was achieved through a process of triangulation, whereby neighboring data points were connected to model the spatial structure of the monitoring network.

The convex hull is defined as the minimal convex polygon that contains all points in a given set. A convex polygon is defined as such if all interior angles are strictly less than 180 degrees~\cite{12a}. These geometric concepts provide fundamental structures for analyzing spatial relationships in various data sets.

\subsection{Temporal Graph Connectivity}

The temporal evolution of the set of points over a 22-year period requires a complex approach to connecting the graphs. While individual graphs are constructed for each month, maintaining the temporal sequence over such an extensive period requires the integration of these distinct monthly graph instances into a unified structure.
A common, though sub-optimal, methodology for achieving this objective entails the random selection of a predetermined number of nodes (e.g., n nodes from the first graph and m nodes from the second graph) and the establishment of $k$ edges between them. This process is iterated sequentially until the final month. However, this arbitrary connection strategy is incapable of leveraging the inherent spatial relationships between observations over time. The modified strategy for temporal graph merging that has been developed overcomes this limitation. In contrast to the conventional approach of establishing random connections, our methodology involves the creation of temporal edges between consecutive monthly graphs. This is achieved by identifying $k$ nearest neighbors based on spatial proximity. Specifically, for each node in the graph of a given month, its $k$ nearest spatial neighbors in the graph of the subsequent month are identified and linked. This methodological approach ensures that temporal connections reflect meaningful spatial similarities, thereby preserving the underlying spatiotemporal structure of the data.  
To ensure the robustness of model evaluation and the prevention of data leakage, the entire dataset is divided into three distinct temporal periods. The training graph is 240 months in duration, followed by 12 months for the validation graph, and the final 12 months for the testing graph. The integration of temporal relationships through these newly added edges effectively overcomes the inherent discontinuities between individual monthly graphs. This allows for the utilization of static graph neural networks in the analysis of dynamic spatio-temporal patterns.

\subsection{Graph Neural Network Models}

In this study, we employ several Graph Neural Network (GNN) architectures to predict dust source emissions based on the constructed proximity graphs. GNNs are particularly well-suited for this task because they can learn node representations by aggregating information from neighboring nodes and edges, effectively capturing complex spatial and temporal dependencies in graph-structured data. The following subsections briefly describe the models used: the Graph Convolutional Network (GCN), GraphSAGE, and Graph Attention Network (GAT), highlighting their key mechanisms and relevance to our problem.

\subsubsection{Graph Convolutional Network (GCN)}

The Graph Convolutional Network (GCN) is a particular type of deep learning model that has been shown to be effective in processing graph-structured data~\cite{15a}.

The GCN is a foundational architecture in the domain of Graph Neural Networks. It is designed to perform feature learning on graph-structured data. Generalization of the convolution concept from grid-like data (e.g., images) to arbitrary graph topologies is achieved by GCNs. The operation of a GCN layer entails the aggregation of feature information from a node's immediate neighbors and its own features. This is followed by a linear transformation and a non-linear activation.

A fundamental aspect of GCNs pertains to their message-passing paradigm, wherein each node undergoes an iterative process of updating its representation. This process entails the combination of its current features with the aggregated features of its direct neighbors. The layer-wise propagation rule for a GCN is commonly formulated as:

 \begin{equation} \label{eq:eq1} 
H^{(k+1))}=\sigma(D\tilde{}^{-\frac{1}{2}}A\tilde{}D\tilde{}^{-\frac{1}{2}}H^{(k)}W^{(k)})\end{equation} 

In this case, $H^{(k))}$ denotes the node feature matrix at layer $k$, $A\tilde{}= A+I$ is the adjacency matrix $A$ with added self-loops (identity matrix $I$), $D\tilde{}$ is the diagonal degree matrix of $A\tilde{}$, $W^{(k)}$ is the trainable weight matrix for layer $k$, and $\sigma$ is an activation function. This process implicitly smooths node features across the graph, thereby allowing nodes to learn representations that incorporate information from their local connectivity patterns. Standard GCNs are frequently regarded as transductive models, as they generally necessitate the complete graph structure to be known during training in order to compute the normalized adjacency matrix.

\subsubsection{Graph Attention Network (GAT)}

The Graph Attention Network (GAT) introduces an attention mechanism into the process of learning node representations on graphs, offering a more flexible and powerful aggregation scheme compared to fixed-weight approaches like standard GCNs~\cite{16a}. GATs are a data-driven approach that allows for the learning of differential importance (attention weights) for each of a node's neighbors during the aggregation process. This is in contrast to the conventional approach of assigning uniform weights to all neighbors.

For a given node $i$ and its neighbor $j$, the unnormalized attention coefficient $e_{ij}$ is computed based on their features: 
\begin{equation} \label{eq:eq2} 
e_{ij}= LeakyReLU(a^{T}.\left[ Wh_{i}\parallel Wh_{i} \right]) \end{equation} 

In this equation, $h_i$ and $h_j$ represent the feature vectors of nodes $i$ and $j$, respectively. $W$ denotes a shared linear transformation weight matrix, 1 is a trainable attention vector, and $\parallel $ indicates concatenation. These unnormalized coefficients are subsequently normalized across all neighbors of node $i$ using a softmax function to obtain the attention weights $\alpha_{ij}$:
 \begin{equation}\alpha_{ij}=\frac{exp(e_{ij})}{\sum_{k\in N(i)}^{}exp(e_{ik})}\label{eq:eq3} 
 \end{equation} 

The node's new representation $h^{'}_{i}$ is then formed as a weighted sum of its neighbors' features, using these learned attention coefficients:

\begin{equation} \label{eq:eq4} 
h^{'}_{i}=\sigma(\sum_{j\in N(i)}^{}a_{ij}Wh_{j})\end{equation} 

This attention mechanism allows GATs to focus on more relevant neighbors, enhancing the model's ability to capture complex relationships and make them inherently inductive, as the attention mechanism can be applied to new, unseen nodes.

\subsubsection{GraphSAGE}

GraphSAGE is a robust and adaptable deep learning framework specifically engineered for inductive learning on large-scale graphs~\cite{14a}. The fundamental operational principle of this system is the generation of novel node representations. These representations are produced through a process of sampling and aggregation of feature information from the node's local neighborhood. This mechanism enables GraphSAGE to effectively generalize to unseen nodes and entire graph structures, a critical capability for inductive inference. The architecture is composed of two interconnected components: an aggregation function and a node update function. 

The aggregation function is responsible for the compilation of information from a node's immediate neighbors. The general form for the aggregated representation $h_k^N(v)$ of node $v$'s neighbors at layer $k$ is given by:

\begin{equation} \label{eq:eq5} 
h^{N}_{k}(v)= AGGREGATE_{k}({h^{u}_{k-1} for u\in N(v)}) \end{equation} 

Common aggregation strategies include:

\begin{itemize}
\item The Mean Aggregator is a statistical measure of central tendency used to calculate the mean value of a set of data. The element-wise average of the neighbor feature vectors is calculated.
\item The LSTM Aggregator is a sequence of neighbor representations that is processed using a Long Short-Term Memory (LSTM) network. The objective of this process is to capture sequential dependencies.
\item The term "pooling aggregator" refers to the application of an operation to neighbor representations. Examples of such operations include element-wise maximum, minimum, or sum.
 \end{itemize}
 
Subsequent to aggregation, the node update function integrates the aggregated neighbor information with the node's own representation from the previous layer to produce its refined representation for the current layer. This phenomenon is commonly articulated as follows:

\begin{equation} \label{eq:eq6} 
h^{v}_{k}= \sigma(W_{k}.CONCAT(h^{v}_{k-1}.h^{N}_{k}(v))+b_{k})\end{equation} 

In this context, $h_k^v$ denotes the new representation of node $v$ at layer $k$. The matrices $W_k$ and $b_k$ are trainable weight matrices and bias vectors, respectively. The symbol . denotes a non-linear activation function (e.g., ReLU). The symbol CONCAT refers to the concatenation operation of the node's own representation and the aggregated neighbor representation.

\section{Data and Experiments}

\subsection{Data}
The Middle East, with its arid and semi-arid climate and minimal rainfall, faces substantial environmental and climatic challenges, particularly in the form of dust storms. These events have been identified as significant contributors to global dust emissions. A significant source of these dust storms in the region is the Tigris and Euphrates basin (TEB), which encompasses an area exceeding 700,000 square kilometers. Within this extensive basin, numerous dust sources are geographically dispersed, and their activity is strongly shaped by a complex interaction of both natural processes and human-induced changes~\cite{19}.

A preliminary investigation revealed Dust Source emissions within the TEB from 2000 to 2021. The identification process, which is detailed in~\cite{20,8}, utilized visual interpretation keys (including shape, size, pattern, tone, texture, shadow, site, and association) applied to MODIS Terra/Aqua satellite imagery. A total of 11,618 dust sources were identified during this period. Specifically, 10,422 of these dust sources (from 2000-2020) served as the training dataset, while the remaining dust sources identified for 2021 were used as test data. For the purpose of analysis, these dust sources are represented as points with specific spatial coordinates.

In this study, the impact of dust emissions is found to depend on several key environmental factors that have been identified in previous research~\cite{9,3a,21}. Vegetation, the presence of vegetation has been shown to stabilize soil, thereby reducing the effects of wind erosion~\cite{14}. Precipitation, an increase in precipitation leads to an increase in soil moisture, which in turn binds particles. Inadequate rainfall can result in dry, erodible soil. Soil Moisture, has been demonstrated to directly affect particle cohesion~\cite{16}. The primary driver is wind speed; high speeds result in the detachment and transportation of particles. Soil temperature has been demonstrated to exert a significant influence on soil moisture through the process of evaporation~\cite{15}. This phenomenon directly impacts erodibility. The erodibility of soil is contingent upon its texture, which can be categorized as either silt, sand, or clay~\cite{7a}. The presence of deeper, unconsolidated layers in the soil profile is indicative of a greater abundance of erodible material, which is measured in terms of soil thickness~\cite{15}. Altitude is a factor influencing local climate, wind patterns, and exposure~\cite{16}. Slope is a factor that exerts influence on two phenomena: water runoff and localized wind effects on soil stability~\cite{16}. The interplay among these factors contributes to the determination of dust source potential and emission intensity.

\subsection{Experiments and results}

The performance of several models including GraphSAGE, GCN, and GAT are investigated on the provided dataset. A series of experiments was conducted, employing various parameter optimization techniques to identify the most effective configurations for each model. To validate the stability of our outcomes, all experiments were independently executed 50 times using different initialization. The models were evaluated against both the reduced and up-sampled data subsets.\\
The primary parameters that were fine-tuned during this research, along with their defined search spaces, include:
\begin{itemize}
    \item Repeat: number of repeats for the whole process (10,
50)
    \item GCN, GAT and Sage-hidden-channel: number of hidden channels and embedding between layers (8, 16, 32, 64, 128,
256, 512)
    \item GCN, GAT and Sage-learning-rate: learning rate of the
GraphSAGE model (0.1, 0.3, 0.5, 0.01, 0.05, 0.001,
0.003)
    \item Epochs: number of max epochs (500, 1000, 2000)
    \item Patience: number of epochs to wait in loss in
validation is not decreased (100, 200)
    \item Criterion: binary cross entropy loss
(BCEWithLogitsLoss)
    \item Optimizer: Adaptative Moment Estimation (Adam) 
\end{itemize}
\subsubsection{GraphSage Results}
As illustrated in Table \ref{tab:combined_performance}, a comparative analysis of various proximity graph structures (Delaunay, Gabriel, Yao$_{r}$, $k$-Nearest Neighbor and Random Graphs) is presented on the performance of a GraphSAGE model. In this analysis, the graphs are temporally connected using a 3-Nearest Neighbor (3-NN) approach. The evaluation metrics encompass Accuracy (ACC), Area Under the Receiver Operating Characteristic Curve (AUC), Precision, and Recall.

The results demonstrate variability in model performance based on the underlying graph topology. The Delaunay triangulation exhibited a commendable performance, attaining an accuracy of 0.67 and an AUC of 0.72, accompanied by a high recall of 0.82. This finding suggests that the model is effective in identifying a large proportion of positive instances, likely due to its comprehensive connectivity, which captures broad proximity relationships.

The Gabriel graph displays slightly diminished overall performance, with an accuracy of 0.64 and an AUC of 0.67. It is noteworthy that the proposed method attains a higher precision (0.67) in comparison to Delaunay. However, this is accompanied by a substantial decline in recall (0.55), signifying a propensity for fewer false positives while concomitantly missing more true positives. This finding corresponds to the sparser connectivity characteristic of the Gabriel graph, which retains edges only if the connecting segment's disk is devoid of other points.

The Yao$_{r}$ Graph exhibit intriguing patterns when the "r" parameter is considered. The Yao$_{2}$ Graph and Yao$_{6}$ Graph models both attain the maximum Accuracy (0.68) and competitive AUC (0.70), along with a robust Precision (0.74 and 0.70, respectively). Yao$_{2}$ Graph is noteworthy for its precision, indicating that it establishes highly reliable connections for positive predictions. Yao$_{4}$ Graph demonstrates slightly lower accuracy (0.65) and precision (0.57), yet it exhibits a remarkably high recall (0.84), thereby surpassing Delaunay in its capacity to capture positive cases, though with a higher rate of false positives. This underscores the trade-off between precision and recall, as well as the impact of angular resolution on graph structure.

It has been demonstrated that, among the $k$-NN graphs, performance is generally observed to increase with 'k'. The 1-NN graph demonstrates the lowest values for both Accuracy (0.54) and Recall (0.33), suggesting a graph with significant sparsity and the potential for disconnection, which hinders the capture of complex patterns. As the value of 'k' increases from 2-NN to 3-NN, there is a corresponding enhancement in both accuracy and recall, with values of 0.63 and 0.61 for 2-NN, and 0.63 and 0.57 for 3-NN, respectively. Notably, 3-NN attained the highest precision (0.75) among all the graph types that were examined, indicating that establishing connections to the three nearest neighbors results in highly accurate positive predictions, despite the possibility of missing some true positives (lower recall in comparison to Delaunay or Yao$_{4}$ Graph).

A random graph is a graph in which connections between nodes are established randomly. It is noteworthy that the number of edges in such a random graph is equivalent to that of a Delaunay graph. It functions as a foundational framework for evaluating the efficacy of structured graph constructions.

The Random Graph, characterized by its arbitrary connections, functions as a critical control in this evaluation. Its consistently lower performance across key metrics (AUC 0.61, ACC 0.59, Precision 0.52, Recall 0.58) empirically substantiates the necessity of integrating geometric or proximity-based information through structured graph constructions to achieve effective GraphSAGE model performance.

\subsubsection{GAT Results}
The table \ref{tab:combined_performance} evaluates the performance of a GAT model, utilizing 3-NN for temporal graph connectivity, across various spatial graph structures.

Specifically, the Delaunay graph consistently achieves the highest AUC of 0.73 and accuracy of 0.65, indicating superior overall discriminative power and correct classification rates. In contrast, the 1-NN graph demonstrates the highest recall rate (0.83), indicating its capacity to effectively identify positive instances. However, its performance in other metrics is comparatively lower. A trade-off between precision and recall is evident, as methods that optimize for one often show a reduction in the other. The Random Graph serves as a baseline, demonstrating the lowest performance across all metrics. This underscores the critical importance of structured proximity graph construction in enhancing the efficacy of GAT models for classification tasks.

\subsubsection{GCN Results}
As illustrated in Table \ref{tab:combined_performance}, a GCN model employing 3-NN for temporal graph connectivity is demonstrated across a range of spatial graph structures. The 1-NN demonstrates the highest AUC of 0.72 and an exceptionally high Recall of 0.96, indicating its superior ability to detect positive instances. The model's accuracy (0.57) and precision (0.53) are moderate.

The Delaunay triangulation exhibited a noteworthy recall rate of 0.94 and an AUC of 0.71, with an accuracy of 0.58. Among the Yao$_{r}$ Graphs, Yao$_{4}$ Graph and Yao$_{6}$ Graph achieve the highest Precision (0.77), suggesting a lower false positive rate, though their Recall is lower (0.56 and 0.58, respectively). Yao$_{2}$ Graph offers a more balanced performance with an Accuracy of 0.61 and an AUC of 0.70. The Gabriel graph demonstrates a relatively balanced performance, with an accuracy of 0.62 and a precision of 0.61. It is important to note that the 3-NN spatial graph data is not available for this GCN configuration.

In summary, the GCN model with 3-NN temporal connectivity demonstrates notable strengths in recall and AUC performance when utilizing 1-NN, thereby underscoring its efficacy in comprehensive detection tasks. The Yao$_{4}$ Graph and Yao$_{6}$ Graph prioritize precision, a desirable characteristic in scenarios where minimizing false alarms is paramount. The selection of the optimal graph is contingent upon the priorities established by the specific application for dust emission forecasting.

The Random Graph serves as a crucial baseline, exhibiting the lowest performance across all metrics (AUC 0.60, ACC 0.56, Precision 0.51, Recall 0.55). This finding underscores the critical importance of incorporating geometric or proximity-based information through structured graph constructions in the GCN model, as random connections yield substantially inferior results.

	\begin{table}[h!]
	\centering
	\caption{Comparative Performance Metrics of LSTM and Graph Neural Network Models on Proximity Graphs with Temporal Graph Connection with $k=3$.)}
	\label{tab:combined_performance}
	\begin{tabular}{l l S[table-format=1.2] S[table-format=1.2] S[table-format=1.2] S[table-format=1.2]}
		\toprule
		\textbf{GNN Model} & \textbf{Input Graph} & \textbf{AUC} & \textbf{ACC} & \textbf{Precision} & \textbf{Recall} \\
		\midrule
		\multirow{8}{*}{\textbf{GraphSAGE}} & Delaunay Graph    & 0.72 & 0.67 & 0.63 & 0.82 \\
		& Gabriel  Graph    & 0.67 & 0.64 & 0.67 & 0.55 \\
		& Yao$_2$ Graph  & 0.70 & 0.68 & 0.74 & 0.66 \\
		& Yao$_4$ Graph  & 0.71 & 0.65 & 0.57 & 0.84 \\
		& Yao$_6$ Graph  & 0.70 & 0.68 & 0.70 & 0.66 \\
		& 1-NN         & 0.65 & 0.54 & 0.73 & 0.33 \\
		& 2-NN         & 0.65 & 0.63 & 0.62 & 0.61 \\
		& 3-NN         & 0.60 & 0.63 & 0.75 & 0.57 \\
        & Random Graph & 0.61 & 0.59 & 0.52 & 0.58 \\
		\midrule
		\multirow{8}{*}{\textbf{GAT}}       & Delaunay  Graph   & 0.73 & 0.65 & 0.56 & 0.73 \\
		& Gabriel  Graph    & 0.62 & 0.53 & 0.54 & 0.77 \\
		& Yao$_2$ Graph  & 0.65 & 0.56 & 0.55 & 0.78 \\
		& Yao$_4$ Graph  & 0.63 & 0.54 & 0.56 & 0.79 \\
		& Yao$_6$ Graph  & 0.61 & 0.61 & 0.57 & 0.80 \\
		& 1-NN         & 0.65 & 0.55 & 0.56 & 0.83 \\
		& 2-NN         & 0.64 & 0.59 & 0.57 & 0.78 \\
		& 3-NN         & 0.62 & 0.52 & 0.50 & 0.77 \\
        & Random Graph & 0.62 & 0.57 & 0.51 & 0.58 \\
		\midrule
		\multirow{8}{*}{\textbf{GCN}}       & Delaunay  Graph   & 0.71 & 0.58 & 0.54 & 0.94 \\
		& Gabriel   Graph   & 0.70 & 0.62 & 0.61 & 0.78 \\
		& Yao$_2$ Graph  & 0.70 & 0.61 & 0.57 & 0.77 \\
		& Yao$_4$ Graph  & 0.69 & 0.58 & 0.77 & 0.56 \\
		& Yao$_6$ Graph  & 0.68 & 0.61 & 0.77 & 0.58 \\
		& 1-NN         & 0.72 & 0.57 & 0.53 & 0.96 \\
		& 2-NN         & 0.70 & 0.59 & 0.57 & 0.78 \\
        & Random Graph & 0.60 & 0.56 & 0.51 & 0.55 \\
		\midrule
		\multirow{1}{*}{\textbf{LSTM}}      & (Not Applicable) & 0.52 & 0.57 & 0.23 & 0.62 \\ % Placeholder for LSTM data
		\bottomrule
	\end{tabular}
\end{table}

\subsubsection{LSTM Results}
Long Short-Term Memory (LSTM) networks, a specialized type of Recurrent Neural Network (RNN), are widely recognized and frequently employed for analyzing sequential data and modeling time series due to their inherent ability to capture long-range dependencies~\cite{17a}. Given the established efficacy of LSTMs in temporal pattern recognition, an LSTM model was utilized as a baseline for comparative analysis in this study. While LSTMs demonstrate robustness for purely temporal sequences, their performance in directly modeling complex spatio-temporal relationships, such as those found in dust source emission dynamics, often presents limitations when compared to graph-aware architectures. This comparison underscores the merits of the proposed GNN framework in addressing the intricate spatial and temporal interdependencies of the dust emission phenomenon.
The following is a thorough exposition of the LSTM model, which was utilized for the classification of time series, on the code and data.
\begin{itemize}
    \item LSTM Model Parameters:
    \begin{itemize}
        \item input-dim $=$ 12, Number of features per time step
        \item hidden-dim $=$ 64,128,...256, Size of LSTM hidden layers
        \item num-layers $=$ 2, Number of LSTM layers
        \item output-dim $=$ 1, Binary classification (0 or 1)
        \item dropout $=$ 0.1,...0,5, Dropout rate for regularization
        \item bidirectional $=$True, Use bidirectional LSTM
    \end{itemize}
    \item Training Parameters:
    \begin{itemize}
        \item batch-size $=$32,..., 128, Number of samples per batch
        \item learning-rate $=$ 0.001, 0.01, 0.1, Learning rate for optimizer
        \item epochs $=$ 50,100, Number of training epochs
    \end{itemize}
    \item Data Parameters:
    \begin{itemize}
        \item sequence-length = 3,...,12, Number of months in each sequence
        \item n-repeats = 5, 10, Number of times to repeat training     
    \end{itemize}
\end{itemize}

The result of the LSTM model is presented in Table\ref{tab:combined_performance}, which is for sequence-length of 6 as the best result. As demonstrated by the findings presented in Table \ref{tab:combined_performance}, the performance of the LSTM model was significantly outperformed by that of the various GNN configurations across key metrics. This finding underscores the efficacy of GNNs in leveraging both spatial and temporal information for more accurate dust source emission forecasting. This task is one in which traditional sequential models, such as LSTMs, may fall short due to their limited ability to explicitly model complex graph structures.
    
\section{Conclusion}
This study advances environmental forecasting by demonstrating that graph neural networks (GNNs) constructed on proximity graphs can effectively capture the complex spatiotemporal dynamics of dust emissions. Unlike conventional methods, our framework integrates fine-scale interactions with broader spatial patterns through graph-based modeling—a capability that holds promise for transforming simulations of diverse Earth system processes.

Our results show that the choice of graph topology—whether Delaunay triangulation, $k$-nearest neighbor graph, or alternative structures—plays a pivotal role in predictive performance. This finding emphasizes the need for research into optimal graph construction methods tailored to specific environmental applications. Although our work focused on dust emission forecasting, the same principles can be extended to other spatially driven environmental processes such as urban air pollution and wildfire spread, where understanding spatial relationships is critical.

%\section{Software and data availability}

%\begin{itemize}

%\item Name of the code/library : Dust-source-emission-forecasting-GNN

%\item Contact: Maryam Sanisales (maryamsan73@gmail.com)

%\item Hardware requirements: X86 64-bit CPU (Intel) and 32 GB RAM

%\item Program language: Python 

%\item The source code and dataset used in the current study are available in the Git Hub\footnote{https://github.com/agmlcenter GIS} repository.

%\end{itemize}

%\section{CRediT authorship contribution statement}

%\textbf{Maryam Sanisales:} Writing – original draft, Visualization, Validation, Software, Methodology, Investigation, Formal analysis, Data curation, Conceptualization.

%\textbf{Zahed Rahmati:} Writing – review \& editing, Supervision, Resources, Project administration, Methodology, Conceptualization.

%\textbf{Ali Darvishi Boloorani:} Writing – review \& editing, Supervision, Resources, Project administration, Methodology, Conceptualization.

%\textbf{Ali Vefghi:} Writing – review \& editing, Visualization, Software, Methodology, Conceptualization.

%\section{Declaration of competing interest}

%The authors declare that they have no known competing financial interests or personal relationships that could have appeared to influence the work reported in this paper.

\FloatBarrier

\bibliographystyle{elsarticle-num}

%---------------------------- Appendix -------------------------------

% \newpage

% \section{Available official codes of models}
\end{document}